\documentclass{article}

    \PassOptionsToPackage{numbers, compress}{natbib}

\usepackage[preprint]{neurips_2026}

\usepackage[utf8]{inputenc} 
\usepackage[T1]{fontenc}    
\usepackage{hyperref}       
\usepackage{url}            
\usepackage{booktabs}       
\usepackage{amsfonts}       
\usepackage{nicefrac}       
\usepackage{microtype}      
\usepackage{xcolor}         
\usepackage{pifont}
\usepackage[table]{xcolor}
\usepackage{wrapfig}
\usepackage{graphicx}

\usepackage{multirow}
\usepackage{amsmath}
\usepackage{amsfonts}
\usepackage{subcaption}
\usepackage{booktabs}
\usepackage{makecell}

\title{Tracing the Arrow of Time: Diagnosing Temporal Information Flow in Video-LLMs}

\author{%
  Peitao Han \textsuperscript{1,2,3,*},
  Fei Cheng \textsuperscript{4,*},
  Lis K. Pereira \textsuperscript{1,2,3},
  Qianying Liu \textsuperscript{5},
  Shigeru Kitazawa \textsuperscript{1,2,3} \\
\textsuperscript{1} The University of Osaka  \\ 
\textsuperscript{2} Center for Information and Neural Networks \\
\textsuperscript{3} National Institute of Information and Communications Technology \\
\textsuperscript{4} Kyoto University
\textsuperscript{5} NII LLMC \\
\textsuperscript{*} \texttt{peitao.han.83r@ecs.osaka-u.ac.jp},
\texttt{feicheng@i.kyoto-u.ac.jp}
}

\begin{document}

\maketitle

\begin{abstract}

The Arrow-of-Time (AoT) task, determining whether a video plays forward or backward by recognizing temporal irreversibility, is one humans solve with near-perfect accuracy, yet frontier Video Large Language Models (Video-LLMs) perform only modestly above chance. This gap raises a key question: do visual backbones fail to encode temporal information, or does information bottleneck lie elsewhere in the Video-LLM architecture? We address this question by isolating the vision encoder from the Video-LLM and tracing temporal information across the encoder, projector, and LLM. We find that video-centric encoders with explicit temporal modeling encode strong temporal signals, whereas frame-centric encoders do not. However, when video-centric representations are passed through a standard Video-LLM architecture, performance often collapses, revealing a bottleneck of temporal information flow. We identify projector design as a key factor: Q-Former disrupts temporal information, while a time-preserved MLP projection substantially improves the LLM's access to such information. Our layer-wise analysis further shows temporal representation dynamics across encoder layers. Guided by these findings, we build a Video-LLM with temporal-aware video-centric encoder, time-preserved projector, and AoT supervision, surpassing human performance on AoT$_{PPB}$ with 98.1\% accuracy, and improving broader temporal reasoning tasks by up to 6.0 points on VITATECS-Direction and 1.3 points on TVBench. Our results show that temporal reasoning in Video-LLMs requires both effective temporal encoding and reliable transfer of this information to the LLM. 
Our code is available in  \url{https://github.com/HanPT831/TAT-VQA}.



\end{abstract}

\section{Introduction}


Modern Video-LLMs typically couple a vision encoder with an LLM through a projector. Their visual backbones range from frame-centric encoders, which process videos as individual frame sequences \citep{wang2024qwen2vlenhancingvisionlanguagemodels, bai2025qwen25vltechnicalreport}, to video-centric encoders that explicitly model temporal dynamics across frames \citep{wang2024internvideo2, bardes2024revisiting, assran2025vjepa2selfsupervisedvideo}. These models have achieved promising performance on broad video understanding benchmarks \citep{chen2024internvlscalingvisionfoundation, mangalam2023egoschemadiagnosticbenchmarklongform, xiao2021nextqa, fu2025videomme}.
However, a growing body of work shows that current Video-LLMs remain weak at temporal reasoning, 
a key prerequisite for understanding real-world dynamics \citep{gao-etal-2025-vision, chow2025physbench}.

\begin{figure}[t]
    \centering
    \includegraphics[width=\columnwidth]{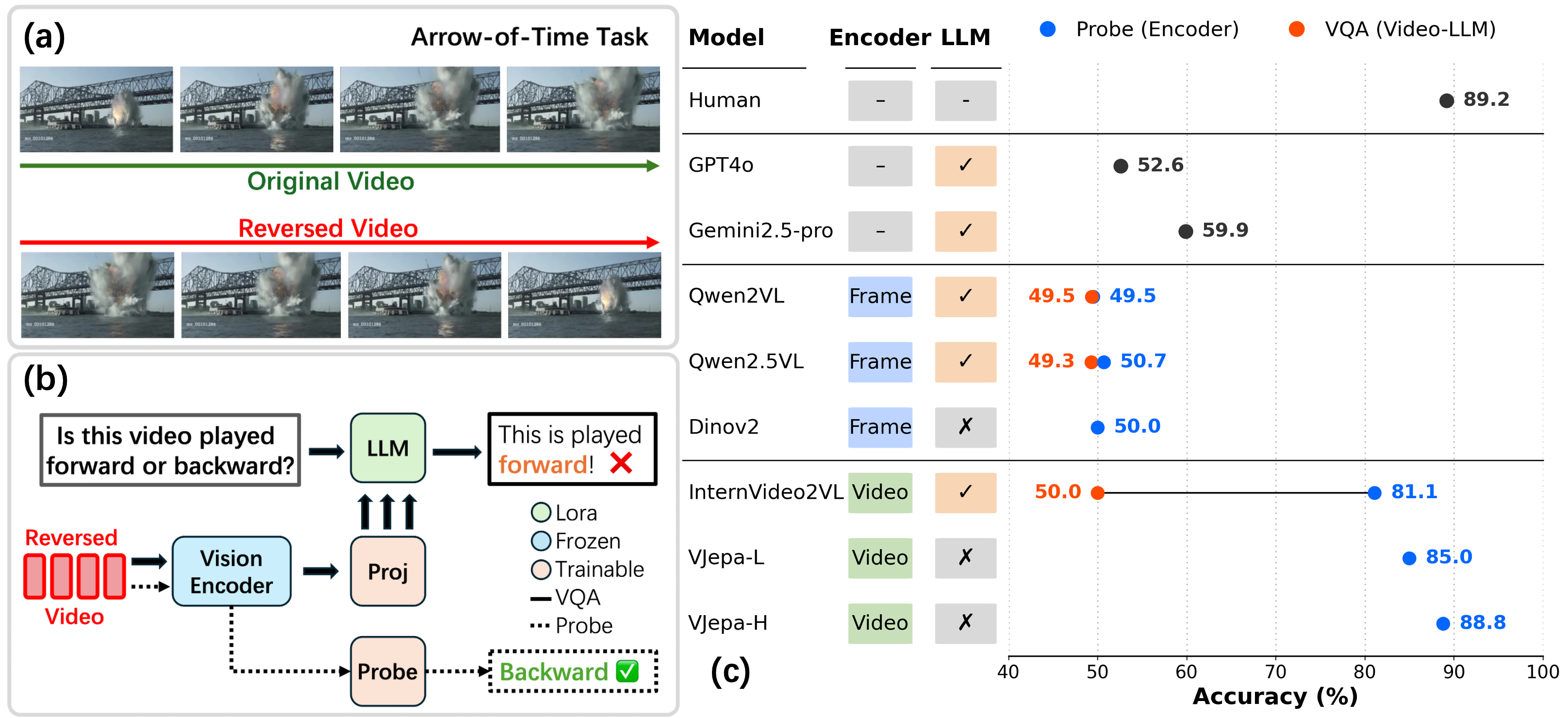}
    \caption{\textbf{(a) The AoT task}: determining whether a video clip is played forward or backward by recognizing physical irreversibility. \textbf{(b) Two AoT evaluation protocols}: probing frozen vision encoders with a lightweight classifier (probe), and fine-tuning Video-LLMs by recasting forward/backward videos as video question answering (VQA) instances. \textbf{(c) AoT performance discrepancy}: video-centric encoders perform strongly on AoT probing, whereas other models remain near chance level.}
    \label{fig:main}
\end{figure}

A particularly striking instance of this weakness arises in the \textit{Arrow of Time} (AoT), the implicit assumption that events unfold irreversibly from past to future, constrained by gravity, entropy, and causality. The AoT \textit{task} operationalizes this principle as a binary judgment: given a short video, determine whether it is playing forward or backward (Figure~\ref{fig:main}(a)). Because the task directly probes physical irreversibility and causal asymmetry, it serves as a clean diagnostic for whether models truly capture temporal dynamics or rely instead on static visual semantics.
Humans solve this task effortlessly \citep{hanyu2023ready}, yet frontier Video-LLMs still fall far short of human-level performance on it \citep{xue2025seeing, matta2025way}.
This substantial gap motivates us to move beyond end-to-end evaluation and trace how temporal information flows across model components. We begin by probing vision encoders to address \textbf{RQ1: can temporal information be encoded by the vision encoder?}

To address this, we cast the AoT task as a binary classification problem using attentive probing \citep{bardes2024revisiting}. Specifically, we categorize vision encoders into two types: (1) \textit{frame-centric encoders}, which primarily encode spatial information from individual frames and rely on temporal positional embeddings for temporal information \citep{wang2024qwen2vlenhancingvisionlanguagemodels, bai2025qwen25vltechnicalreport, oquab2023dinov2}, (2) \textit{video-centric encoders}, which explicitly model temporal dynamics across multiple frames \citep{tong2022videomae, wang2024internvideo2, bardes2024revisiting}.
Our results reveal that temporal information can be successfully encoded by video-centric encoders utilizing explicit temporal modeling, whereas frame-centric encoders fail catastrophically (Figure~\ref{fig:main} (c)).
This contrast suggests that explicit temporal modeling is critical for AoT judgment.
Moreover, the poor VQA performance of Video-LLMs, despite fine-tuning on the same AoT data, points to an overlooked bottleneck: temporal information can be encoded by video-centric encoders, yet still fails to be effectively transferred to the language model.
Rather than improving frame-centric encoders with additional temporal modules \citep{rasekh2026enhancing} or reinforcement learning \citep{xue2025seeing}, we focus on \textbf{RQ2: how can temporal information be effectively transferred from video-centric encoders to LLMs?}

\begin{figure}[t]
    \centering
    \includegraphics[width=\columnwidth]{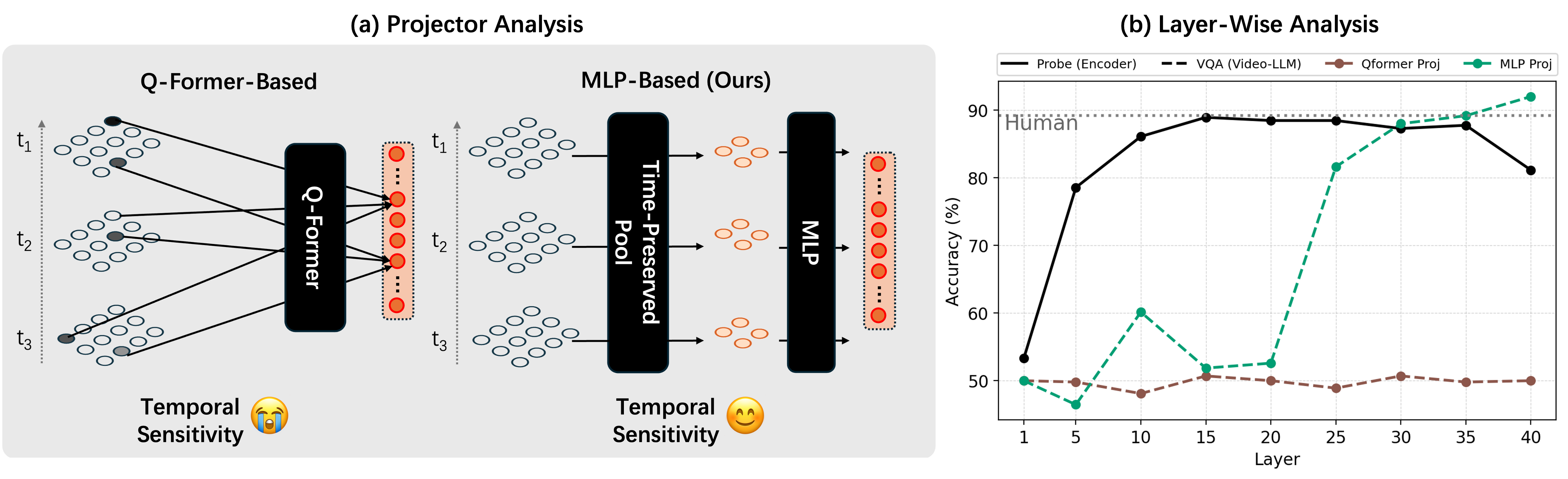}
    \caption{\textbf{(a) Projector analysis}: We compare two projector designs: Q-Former and MLP projector. The MLP projector preserves temporal sensitivity across the projection step, whereas the Q-Former disrupts it. \textbf{(b) Layer-wise analysis}: Representations from different layers of frozen video-centric encoders are used for probing and Video-LLM fine-tuning. Results show that temporal information can be effectively transferred when temporal structure is preserved. 
    }
    \label{fig:projector}
\end{figure}

To this end, we investigate projector design for temporal reasoning by comparing two commonly used architectures: Q-Former and multi-layer perceptron (MLP). 
Our results show that the projector is a key bottleneck in temporal information flow: when the projector (e.g., Q-Former) disrupts the temporal information encoded in video representations, AoT performance in Video-LLMs drops markedly. This may explain why Video-LLMs using Q-Former (i.e., InternVideo2VL) remain largely insensitive to AoT supervision. In contrast, our proposed MLP projector better preserves temporal information and leads to stronger AoT performance (Figure~\ref{fig:projector}(a)).
Our layer-wise analysis (Figure~\ref{fig:projector}(b)) further shows that temporal information is encoded in the early layers of video-centric encoders. In deeper layers, this information can be effectively decoded by LLM when transferred through a time-preserved projector.
Our layer-wise results also show that CLIP-style \citep{radford2021learning} vision-language alignment makes temporal information less transferable to LLMs.



By identifying where temporal recognition emerges and how this information is subsequently transformed, we provide a mechanistic explanation for why current Video-LLMs fail on AoT.
Based on these findings, we pinpoint the key bottlenecks in temporal information transfer and surpass human performance on AoT$_{PPB}$ \citep{matta2025way}, a psychophysics-grounded AoT benchmark, by combining a temporal-aware video encoder and a time-preserved projector with LLMs. More importantly, we investigate \textbf{RQ3: Can AoT supervision improve Video-LLM performance on broader temporal reasoning tasks?} Our results show that AoT supervision serves as an effective training signal beyond the task itself, improving general temporal reasoning in Video-LLMs and yielding gains of up to 6.0 points on VITATECS-Direction \citep{li2024vitatecsdiagnosticdatasettemporal} and 1.3 points on TVBench \citep{cores2025tvbench}.

In summary, we (i) show that video-centric encoder with explicit temporal modeling is critical for encoding temporal information; (ii) identify projector design and CLIP-style training as key bottlenecks for transferring temporal information to LLMs; (iii) improve Video-LLM AoT performance from chance level to beyond human accuracy; and (iv) demonstrate that AoT supervision also benefits broader temporal reasoning tasks.






\section{Related Work}


\noindent\textbf{Video Large Language Models} Video-LLMs extend LLMs to dynamic visual inputs by coupling visual encoders with language models through projection layers, achieving strong performance on a range of video understanding tasks \citep{chen2024internvlscalingvisionfoundation, wang2023internvid, xiao2021nextqa, fu2025videomme}.
\textit{Frame-centric encoders} process videos as individual frames and rely on temporal positional embeddings or token aggregation to represent temporal information \citep{wang2024qwen2vlenhancingvisionlanguagemodels, bai2025qwen25vltechnicalreport, weng2024longvlmefficientlongvideo, jin2024chatuniviunifiedvisualrepresentation}. In contrast, \textit{video-centric encoders} explicitly model temporal dynamics across frames \citep{tong2022videomae, wang2024internvideo2, chen2026vljepajointembeddingpredictive}. Projectors further map visual features into the LLM embedding space, commonly using Q-Former-based or MLP-based designs \citep{ryoo2025xgenmmvidblip3videoneed32, zhang2023videollamainstructiontunedaudiovisuallanguage, wang2024internvideo2, bai2025qwen25vltechnicalreport, cheng2024videollama2advancingspatialtemporal}. Despite these advances, current Video-LLMs remain limited in temporally sensitive reasoning \citep{xue2025seeing, joseph2026interpretingphysicsvideoworld, shi2025causalitymatterstemporalinformation, cores2025tvbench}. Our work studies this limitation by tracing how temporal information is encoded in visual backbones and transferred through projectors to the LLM.


\noindent\textbf{Temporal Reasoning in Video-LLMs} Recent benchmarks increasingly evaluate temporal reasoning in Video-LLMs. General video benchmarks cover common video understanding abilities \citep{fu2025videomme, li2024mvbenchcomprehensivemultimodalvideo, mangalam2023egoschemadiagnosticbenchmarklongform, xiao2021nextqa}, but several studies have shown that they offer limited sensitivity \citep{cores2025tvbench, xue2025seeing, buch2022revisiting}. To address this, temporal benchmarks have been proposed to test abilities such as temporal localization, physical commonsense, and causality \citep{cores2025tvbench, li2024vitatecsdiagnosticdatasettemporal, matta2025way, xue2025seeing, bordes2025intphys2benchmarkingintuitive}. Among these, the AoT task \citep{hanyu2023ready, xue2025seeing, pickup2014seeing, wei2018learning} provides a direct diagnostic for temporal reasoning: a model must determine whether a video is played forward or backward by recognizing temporal irreversibility. Yet frontier Video-LLMs remain far below human performance \citep{matta2025way}. We use AoT as a diagnostic lens to trace how temporal information flows in Video-LLMs and further show that AoT supervision improves broader temporal reasoning.

\section{Preliminaries}

\subsection{The Arrow-of-Time (AoT) Task}
The AoT task assesses whether a model can infer the temporal direction of a video, i.e., whether a clip is played forward (\texttt{F}) or backward (\texttt{B}). Formally, let a forward video clip be a sequence of $T$ frames
$\mathbf{v} = (\mathbf{x}_1,\ldots,\mathbf{x}_T)$, where $\mathbf{x}_n\in \mathbb{R}^{H \times W \times C}$, with $H$ and $W$ denoting the spatial resolution, and $C$ the number of channels. Given a time-reversal operation
$\mathcal{R}(\mathbf{v})  = (\mathbf{x}_T,\ldots,\mathbf{x}_1)$,
we create forward/backward samples as follows
\begin{equation}
\tilde{\mathbf{v}}=
\begin{cases}
\mathbf{v}, & y=\texttt{F},\\
\mathcal{R}(\mathbf{v}), & y=\texttt{B},
\end{cases}
\end{equation}
where $y$ denotes the direction of the observed clip $\tilde{\mathbf{v}}$.


\subsection{Video Large Language Models (Video-LLM)}

A typical Video-LLM consists of three components: a vision encoder, a projector, and an LLM. Given an input video, the vision encoder extracts visual representations, the projector maps these representations into visual tokens compatible with the LLM, and the LLM generates responses conditioned on both the visual tokens and the text instructions.

\noindent\textbf{Vision encoder.}
We distinguish between two types of vision encoders. \textit{Frame-centric encoders} process videos mainly as individual frames and rely on additional mechanisms for temporal modeling. \textit{Video-centric encoders}, in contrast, explicitly model temporal dynamics across frames.

\noindent\textbf{Projector.}
The projector compresses the visual representations into a smaller number of visual tokens and maps them into the LLM embedding space. 

In this work, we investigate which configuration of vision encoder and projector design best encodes and transfers temporal information.

\subsection{Training Stages of Video-LLMs}

Video-LLMs are typically developed through three sequential training stages.

\noindent\textbf{$Stage1$: Vision-only Pretraining.}
The vision encoder is first trained on image or video data using vision-only objectives, learning visual representations solely from visual signals.

\noindent\textbf{$Stage2$: Vision-Language Alignment.}
The pretrained vision encoder is then aligned with language to reduce the modality gap, typically through CLIP-style training \citep{wang2023internvid, bolya2026perception, wang2024internvideo2}, making visual representations more compatible with language models.

\noindent\textbf{Instruction Tuning.}
Finally, the vision encoder is connected to an LLM through a projector and fine-tuned on multimodal instruction-tuning data. This stage produces the complete Video-LLM and enables video-conditioned language generation.
\section{RQ1: Can Temporal Information Be Encoded by the Vision Encoder?}
\label{section:rq1}

\subsection{Probing Vision Encoders}

To assess whether temporal information is encoded in vision encoders, 
we train an attentive probe classifier on top of frozen vision encoders. The probe consists of an attention-based pooling layer followed by an MLP classifier.
Given a video clip $\tilde{\mathbf{v}} \in \mathbb{R}^{T \times H \times W \times C}$, a frozen vision encoder maps the input video $\tilde{\mathbf{v}}$ to visual representations $\mathbf{H} \in \mathbb{R}^{T' \times H' \times W' \times d}$, where $T'=T/\tau$, $H'=H/p$, and $W'=W/p$. Here, $\tau$ denotes the temporal tubelet size, $p$ denotes the spatial patch size, and $d$ is the hidden dimension. We flatten the spatiotemporal dimensions of $\mathbf{H}$ into a token sequence $\mathbf{H} \in \mathbb{R}^{N \times d}$, where $N=T'H'W'$. Given a learnable query token $\mathbf{q} \in \mathbb{R}^{d}$, attentive probing computes $\tilde{\mathbf{q}} = \mathrm{softmax}\!\left((\mathbf{q}\mathbf{W}_Q)(\mathbf{H}\mathbf{W}_K)^\top / \sqrt{d}\right)(\mathbf{H}\mathbf{W}_V)$, followed by $p(y \mid \tilde{\mathbf{v}}) = \mathrm{softmax}(\mathrm{MLP}(\tilde{\mathbf{q}}))$, where $y \in \{\texttt{F}, \texttt{B}\}$. Here, $\mathbf{W}_Q$, $\mathbf{W}_K$, and $\mathbf{W}_V$ are learnable projection matrices.

\subsection{Experiments}
\noindent\textbf{Models.}
For \textit{video-centric encoders}, we evaluate VJepa-L\&H \citep{bardes2024revisiting}, and 
InternVideo2-1B$_{stage1}$ and InternVideo2-1B$_{stage2}$ \citep{wang2024internvideo2}.
For \textit{frame-centric encoders}, we evaluate Qwen2VL$_{\mathrm{ViT}}$ \citep{wang2024qwen2vlenhancingvisionlanguagemodels}, Qwen2.5VL$_{\mathrm{ViT}}$ \citep{bai2025qwen25vltechnicalreport} and DINOv2-1B \citep{oquab2023dinov2}.
For a fair comparison with probing, we additionally report the VQA performance of InternVideo2VL-8B \citep{wang2024internvideo2}, Qwen2VL-7B \citep{wang2024qwen2vlenhancingvisionlanguagemodels}, and Qwen2.5VL-7B \citep{bai2025qwen25vltechnicalreport} before and after fine-tuning on AoT VQA data constructed from the same videos used for probe training. 
During fine-tuning, we keep the vision encoder frozen, train the projector, and update the LLM with LoRA \citep{hu2021loralowrankadaptationlarge}. The differences between InternVideo2 models are provided in Appendix~\ref{sec:app_internvideo}.

\noindent\textbf{Training Data.}
We use videos from the training split of Something-Something v2 (SSv2) \citep{goyal2017something}.
To construct balanced forward/backward supervision, we randomly reverse 50\% of the videos during training. Videos are resized to $224\times 224$ resolution, and we uniformly sample 16 frames per video clip as input. Other training details are given in Appendix~\ref{sec:app_experiments}. For Video-LLM fine-tuning, forward/backward videos are converted into VQA-style samples using manually crafted templates, which are provided in Appendix~\ref{sec:aot_vqa_templates}.

\noindent\textbf{AoT Evaluation Data.}
We evaluate trained probes and Video-LLMs on the psychophysics-grounded Arrow-of-Time benchmark ($AoT_{PPB}$) \citep{matta2025way}. This benchmark consists of videos depicting physically or causally irreversible events and enables direct comparison between human and model performance. Each video is presented in both forward and reversed order, and we report overall accuracy.

\subsection{Results}


        
    

    

  
\begin{table*}[t]
  \small
  \centering
  \caption{\label{table:rq1_results}
  \textbf{Accuracy on $AoT_{PPB}$.}
  The AoT column indicates whether the model is trained on AoT task.
  The Encoder column denotes the encoder type: video-centric encoders and frame-centric encoders.
  Results for GPT-4o and Gemini2.5-pro are taken from \citet{matta2025way}.
  }

  \begin{minipage}[t]{0.48\textwidth}
  \centering
  \textbf{Probing}
  \vspace{0.3em}

  \begin{tabular}{lccc}
    \toprule
    \textbf{Model} & \textbf{AoT} & \textbf{Encoder} & \textbf{Acc.} \\
    \midrule
    Human & $-$ & $-$ & \textbf{89.2} \\
    \midrule
    Qwen2VL$_{ViT}$      & \ding{51} & Frame & 49.5 \\
    Qwen2.5VL$_{ViT}$    & \ding{51} & Frame & 50.7 \\
    DINOv2               & \ding{51} & Frame & 50.0 \\
    \cmidrule(lr){1-4}
    VJepa-L              & \ding{51} & Video & 85.0 \\
    VJepa-H              & \ding{51} & Video & \textbf{88.8} \\
    InternVideo2$_{stage1}$ & \ding{51} & Video & 81.1 \\
    InternVideo2$_{stage2}$ & \ding{51} & Video & 49.5 \\
    \bottomrule
  \end{tabular}
  \end{minipage}
  \hfill
  \begin{minipage}[t]{0.48\textwidth}
  \centering
  \textbf{VQA}
  \vspace{0.3em}

  \begin{tabular}{lccc}
    \toprule
    \textbf{Model} & \textbf{AoT} & \textbf{Encoder} & \textbf{Acc.} \\
    \midrule
    Human & $-$ & $-$ & \textbf{89.2} \\
    \midrule
    GPT-4o          & \ding{55} & $-$    & 52.6 \\
    Gemini2.5-pro   & \ding{55} & $-$    & \textbf{59.9} \\
    \cmidrule(lr){1-4}
    Qwen2VL-7B      & \ding{55} & Frame & 50.0 \\
    Qwen2.5VL-7B    & \ding{55} & Frame & 49.3 \\
    InternVideo2VL  & \ding{55} & Video & 50.0 \\
    \cmidrule(lr){1-4}
    Qwen2VL-7B      & \ding{51} & Frame & 49.5 \\
    Qwen2.5VL-7B    & \ding{51} & Frame & 49.3 \\
    InternVideo2VL  & \ding{51} & Video & 50.0 \\
    \bottomrule
  \end{tabular}
  \end{minipage}

\end{table*}

Table~\ref{table:rq1_results} summarizes the overall AoT accuracy on AoT$_{PPB}$. In the probing setting, video-centric encoders substantially outperform frame-centric encoders, except for the language-aligned InternVideo2$_{\textit{stage2}}$, which we further analyze in Section~\ref{section:rq2}. VJepa-L, VJepa-H, and InternVideo2$_{\textit{stage1}}$ achieve strong performance, reaching 85.0\%, 88.8\%, and 81.1\% accuracy, respectively. In contrast, frame-centric encoders remain near chance level, with Qwen2VL$_{\mathrm{ViT}}$ and Qwen2.5VL$_{\mathrm{ViT}}$ achieving 49.5\% and 50.7\%, respectively. These results indicate that AoT information can be effectively encoded by video-centric encoders with explicit cross-frame temporal modeling, whereas frame-centric encoders fail to capture such dynamics.

In the VQA setting, Video-LLMs still perform poorly on AoT. Gemini2.5-pro achieves the best performance among Video-LLMs at 59.9\%, but remains far below human performance of 89.2\%, while other Video-LLMs stay close to chance level. Fine-tuning with AoT supervision does not substantially improve performance either. Notably, although the InternVideo2$_{\textit{stage1}}$ encoder achieves 81.1\% accuracy under probing, InternVideo2VL reaches only 50.0\% in the VQA setting. This discrepancy suggests that temporal information encoded in the visual backbone is not effectively transferred to LLM.

Overall, these results support two conclusions. First, temporal information can be effectively encoded in \textit{video-centric encoders} with explicit temporal modeling. Second, current Video-LLMs fail to effectively transfer this information to LLMs, revealing an overlooked bottleneck in temporal information transfer. 
This motivates RQ2: How can temporal information be effectively transferred from video-centric encoders to LLMs?

\section{RQ2: How Can Temporal Information Be Effectively Transferred From  Video-Centric Encoders to LLMs?}
\label{section:rq2}

RQ1 showed that video-centric encoders can effectively encode temporal information, yet this information often fails to transfer effectively to the LLM. This suggests that the key bottleneck lies not only in whether temporal information exists in the visual backbone, but also in how it is transferred to the LLM. In this section, we first compare different projector architectures to investigate how to improve temporal information transfer. We then conduct a layer-wise analysis to assess how language alignment affects temporal information within the video encoder.

\subsection{Projector Design}

The projector maps video representations into a smaller set of visual tokens for the LLM. Since AoT recognition depends on cues of temporal irreversibility, the key distinction among projector designs is whether temporal information is preserved during this transfer. We compare three projector designs.

\noindent\textbf{Q-Former.}
Q-Former \citep{li2023blip2bootstrappinglanguageimagepretraining} uses a fixed number of learnable query tokens that attend to visual representations and the text instruction, producing a compact set of visual tokens. 


\noindent\textbf{Time-Preserved MLP.}
The time-preserved MLP preserves the temporal dimension $T'$ while pooling only the spatial dimensions $H' \times W'$. A two-layer MLP then projects the resulting tokens into the LLM embedding space. Compared with the time-compressed MLP, this design maintains the order of visual evidence across frames, although it overlooks some spatial details.

\noindent\textbf{Time-Compressed MLP.}
The time-compressed MLP pools video representations along the temporal dimension $T'$ and then maps the pooled features into the LLM via a two-layer MLP. This design serves as an ablation that preserves more visual information but downsamples the temporal resolution.

\noindent\textbf{Experiments of Projector Analysis.}
We evaluate projector designs by measuring downstream AoT VQA performance, using InternVideo2$_{\textit{stage1}}$ and InternVideo2$_{\textit{stage2}}$ as video encoders. We compare projectors under two visual-token budgets. All other experimental settings follow RQ1 (Section~\ref{section:rq1}).

\begin{figure}[t]
    \centering
    \includegraphics[width=\linewidth]{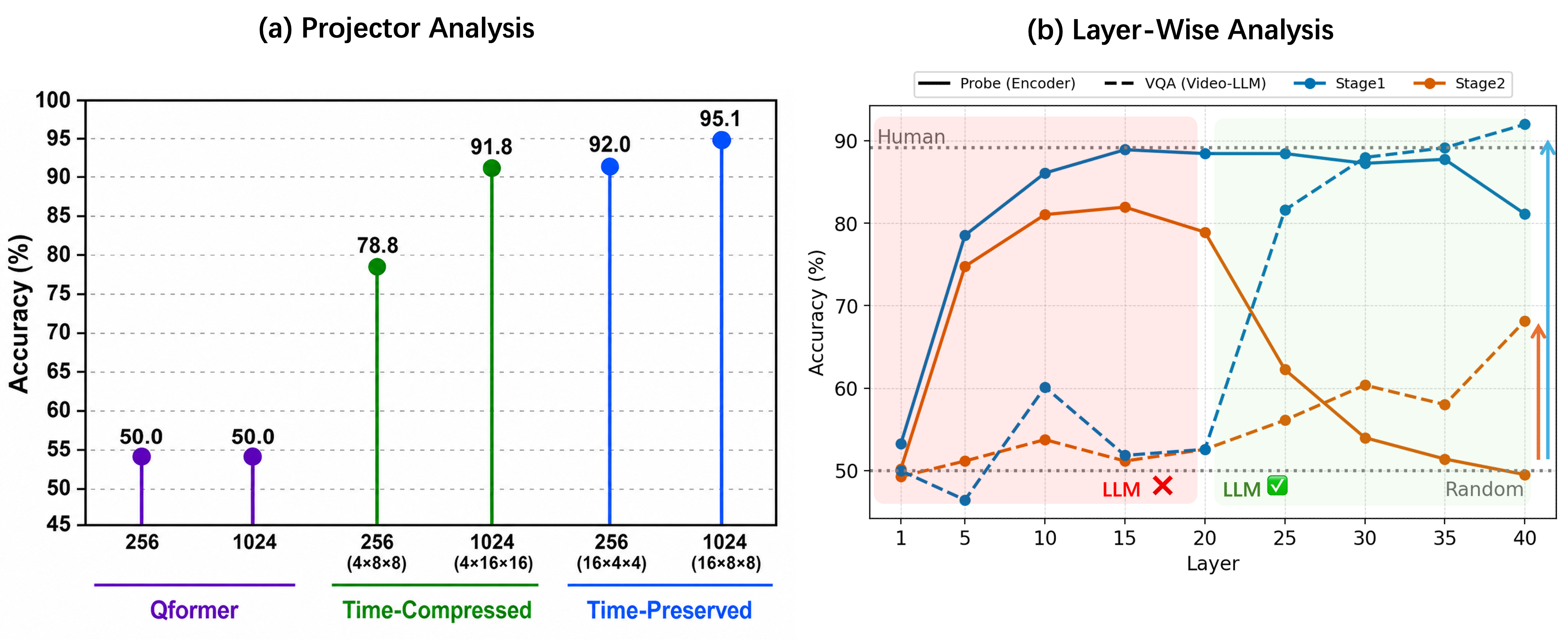}
    \caption{\textbf{(a) Projector Design}: projectors compress video representations $\mathbf{H} \in \mathbb{R}^{T' \times H' \times W' \times d}$ into a smaller set of video tokens. We compare Q-Former, time-compressed MLP and time-preserved MLP. \textbf{(b) Layer-wise analysis:} We investigate the encoder discrepancy between $stage1$ and $stage2$ through layer-wise analysis. Video representations from different layers of the frozen video-centric encoder are used to train the probe and the LLM, respectively. 
    }
    \label{fig:proj_layer}
\end{figure}

\subsection{Results of Projector Design}
Figure~\ref{fig:proj_layer}(a) shows the InternVideo2$_{\textit{stage1}}$ results; complete results for projector design are reported in Appendix~\ref{sec:app_proj}.
Given video representations $T' \times H' \times W' = 16 \times 16 \times 16$, we compare different projector designs under two visual-token budgets: 256 and 1024 tokens. Q-Former remains close to chance-level accuracy across both token budgets, indicating that it severely disrupts temporal information. Prior work has shown that Q-Former can damage spatial information in image understanding \citep{yao2024decodecouplingtokencompression, cha2023honeybee, lin-etal-2024-preserve}; our results extend this observation to video, showing that such projection can also weaken temporal information in video representations. By contrast, MLP-based projectors consistently outperform Q-Former which is the projector of InternVideo2VL (50.0\%), suggesting that a less intrusive projector better preserves the original video representation and facilitates temporal information transfer to the LLM.

Among MLP-based projectors, time-preserved projection consistently outperforms time-compressed projection. Under the 1024-token setting, time-preserved MLP achieves 95.1\% accuracy, compared with 91.8\% for time-compressed MLP. The same trend appears under the 256-token setting: preserving temporal resolution ($16 \times 4 \times 4$) reaches 92.0\%, while the time-compressed setting ($4 \times 8 \times 8$) reaches only 78.8\%. Increasing the token budget from 256 to 1024 consistently improves performance. These results show that AoT-relevant information depends strongly on preserving temporal information during visual-to-language transfer. 
Overall, the projector must preserve temporal information; otherwise, such information encoded in the video representation can be lost.

\subsection{Layer-wise Analysis}
\label{sec:layer_wise}

Table~\ref{table:rq1_results} shows that InternVideo2$_{\textit{stage1}}$ achieves much stronger probing performance than InternVideo2$_{\textit{stage2}}$ (81.1\% vs. 49.5\%). This discrepancy raises a key question: how does language alignment change the accessibility and transferability of temporal information? To answer this, we trace the representation dynamics across layers of both $stage1$ and $stage2$ encoders.

\noindent\textbf{Layer-wise probing.}
For each encoder layer, we train the same attentive probe used in RQ1 on top of frozen encoders to measure how explicitly temporal information can be read out.

\noindent\textbf{Layer-wise fine-tuning.}
For each selected layer, we feed its representations into the time-preserved MLP projector and fine-tune the projector and LLM on AoT VQA data, measuring how effectively this information is transferred to the LLM.

\noindent\textbf{Experiments of Layer-Wise Analysis.}
We analyze InternVideo2 at both training stages, $Stage1$ and $Stage2$, under both layer-wise probing and layer-wise fine-tuning. For layer-wise Video-LLM fine-tuning, we follow InternVideo2VL using Mistral-7B-Instruct as language backbone. We adopt time-preserved MLP projector under a 256-token budget for efficiency. The projector reduces only the spatial resolution from $16 \times 16$ to $4 \times 4$ while keeping the temporal dimension unchanged at $T'=16$.
All other settings follow RQ1 (Section~\ref{section:rq1}). Additional layer-wise results under different settings are provided in Appendix~\ref{sec:app_layer}.

\subsection{Results of Layer-Wise Analysis}
Figure~\ref{fig:proj_layer}(b) shows the layer-wise AoT performance of InternVideo2 before and after language alignment ($Stage1$ and $Stage2$). We highlight two main findings.

\noindent\textbf{Temporal information is effectively encoded in video-centric encoders.}
In the probing setting, both  InternVideo2$_{\textit{stage1}}$ and  InternVideo2$_{\textit{stage2}}$ begin near chance level in the first layer, but their performance rapidly improves from the fifth layer. For  InternVideo2$_{\textit{stage1}}$, probing accuracy rises from around chance level to nearly human-level performance by the 15th layer.  InternVideo2$_{\textit{stage2}}$ exhibits a similar pattern, indicating that AoT-relevant temporal information continues to be encoded even after language alignment. Together, these results confirm that video-centric encoders can encode temporal cues required for AoT task.

\noindent\textbf{LLMs decode temporal information from late layers, and language alignment (i.e., \textit{stage2}) harms temporal understanding.}
The VQA results reveal a complementary trend. For  InternVideo2$_{\textit{stage1}}$, VQA accuracy remains close to chance in early layers but increases sharply in the 25th layer, eventually surpassing human performance. By contrast,  InternVideo2$_{\textit{stage2}}$ yields consistently weaker AoT performance. Specifically, probing accuracy for $Stage2$ drops markedly after the 25th layer and approaches chance level in the final layer. This indicates that the temporal information in  InternVideo2$_{\textit{stage2}}$ is less effectively transferred to the LLM, likely transforming temporal information into a more implicit and language-conditioned representation.


\definecolor{stageone}{RGB}{54,95,145}
\definecolor{stageonebg}{RGB}{235,242,250}
\definecolor{stagetwo}{RGB}{196,114,36}
\definecolor{stagetwobg}{RGB}{252,241,228}
\definecolor{gaincolor}{RGB}{34,139,34}
\definecolor{degradecolor}{RGB}{190,60,60}
\definecolor{neutralcolor}{RGB}{120,120,120}

\begin{table*}[t]
\centering
\caption{\textbf{Main Results.} The $+$AoT column indicates whether AoT data is included during instruction tuning. Colored deltas report performance changes relative to the corresponding counterpart trained without AoT. 
Baseline models are not directly comparable to ours due to substantially larger training data and different experimental settings.}

\label{tab:main}
\setlength{\tabcolsep}{2.5pt}
\resizebox{\textwidth}{!}{
\begin{tabular}{l|c|cc|cccc|cc}
\toprule
\multirow{3}{*}{\textbf{Model}} 
& \multirow{3}{*}{\textbf{$+$AoT}}  
& \multicolumn{6}{c|}{\textbf{Temporal Benchmarks}}
& \multicolumn{2}{c}{\textbf{General Benchmarks}} \\
\cmidrule(lr){3-8} \cmidrule(lr){9-10}
& 
& \multirow{2}{*}{\textbf{AoT$_{PPB}$}} 
& \multirow{2}{*}{\makecell{\textbf{VITATECS}\\direction}}
& \multicolumn{4}{c|}{\textbf{TVBench}} 
& \multirow{2}{*}{\textbf{MVBench}} 
& \multirow{2}{*}{\makecell{\textbf{Video-MME}\\w/o subs}} \\
& 
& 
&
& Averaged
& AS 
& AL 
& MD 
& 
& \\
\midrule
Random  
& $-$ 
& 50.0 
& 50.0
& 33.3 
& 50.0 
& 25.0 
& 25.0 
& 27.3 
& 25.0  \\
\midrule
GPT-4o 
& $-$ 
& 52.6 
& $-$
& 39.9 
& 59.3 
& 25.0 
& 30.6 
& 64.6 
& 71.9  \\
Gemini2.5-pro 
& $-$ 
& 59.9
& $-$ 
& 62.6
& $-$ 
& $-$
& $-$ 
& 70.6 
& $-$  \\
\midrule


InternVideo2VL 
& $-$ 
& 50.0 
& 88.7
& 40.7
& 65.2 
& 38.8 
& 27.0 
& 60.3 
& 41.9  \\

Qwen2VL-7B
& $-$ 
& 50.0 
& 86.6
& 43.8 
& 63.8 
& 41.3 
& 22.4 
& 67.0 
& 63.3  \\

Qwen2.5VL-7B
& $-$ 
& 49.3 
& 80.0
& 45.2 
& 64.8 
& 38.8 
& 33.6
& 69.6
& 65.1  \\
\midrule
\midrule

Mistral-7B \\

\rowcolor{stageonebg}
\qquad+ \textcolor{stageone}{InternVideo2$_{\textit{stage1}}$} 
& \ding{55} 
& 48.6  {\scriptsize \textcolor{neutralcolor}{($-$)}} 
& 88.8  {\scriptsize \textcolor{neutralcolor}{($-$)}} 
& 40.1  {\scriptsize \textcolor{neutralcolor}{($-$)}} 
& 64.6  {\scriptsize \textcolor{neutralcolor}{($-$)}} 
& 34.4  {\scriptsize \textcolor{neutralcolor}{($-$)}} 
& 25.4  {\scriptsize \textcolor{neutralcolor}{($-$)}} 
& 56.3  {\scriptsize \textcolor{neutralcolor}{($-$)}} 
& 41.3  {\scriptsize \textcolor{neutralcolor}{($-$)}} \\

\rowcolor{stageonebg}
\qquad+ \textcolor{stageone}{InternVideo2$_{\textit{stage1}}$} 
& \ding{51} 
& 94.8 {\scriptsize \textcolor{gaincolor}{(+46.2)}} 
& 94.2 {\scriptsize \textcolor{gaincolor}{(+5.4)}} 
& 41.4 {\scriptsize \textcolor{gaincolor}{(+1.3)}} 
& 68.7 {\scriptsize \textcolor{gaincolor}{(+4.1)}} 
& 38.1 {\scriptsize \textcolor{gaincolor}{(+3.7)}} 
& 29.7 {\scriptsize \textcolor{gaincolor}{(+4.3)}} 
& 56.1 {\scriptsize \textcolor{degradecolor}{(-0.2)}} 
& 41.0 {\scriptsize \textcolor{degradecolor}{(-0.3)}}  \\

\rowcolor{stagetwobg}
\qquad+ \textcolor{stagetwo}{InternVideo2$_{\textit{stage2}}$} 
& \ding{55} 
& 50.0  {\scriptsize \textcolor{neutralcolor}{($-$)}}
& 93.1 {\scriptsize \textcolor{neutralcolor}{($-$)}}
& 39.0 {\scriptsize \textcolor{neutralcolor}{($-$)}}
& 64.9 {\scriptsize \textcolor{neutralcolor}{($-$)}}
& 29.4 {\scriptsize \textcolor{neutralcolor}{($-$)}}
& 29.3 {\scriptsize \textcolor{neutralcolor}{($-$)}}
& 55.4 {\scriptsize \textcolor{neutralcolor}{($-$)}}
& 39.5 {\scriptsize \textcolor{neutralcolor}{($-$)}}\\

\rowcolor{stagetwobg}
\qquad+ \textcolor{stagetwo}{InternVideo2$_{\textit{stage2}}$} 
& \ding{51} 
& 79.3 {\scriptsize \textcolor{gaincolor}{(+29.3)}} 
& 93.2 {\scriptsize \textcolor{gaincolor}{(+0.1)}}
& 39.6 {\scriptsize \textcolor{gaincolor}{(+0.6)}}
& 67.3 {\scriptsize \textcolor{gaincolor}{(+2.4)}}
& 31.3 {\scriptsize \textcolor{gaincolor}{(+1.9)}}
& 25.4 {\scriptsize \textcolor{degradecolor}{(-3.9)}}
& 55.4 {\scriptsize \textcolor{neutralcolor}{($-$)}}
& 40.4 {\scriptsize \textcolor{gaincolor}{(+0.9)}}\\

\midrule

Qwen2.5-7B \\

\rowcolor{stageonebg}
\qquad+ \textcolor{stageone}{InternVideo2$_{\textit{stage1}}$}  
& \ding{55} 
& 50.7  {\scriptsize \textcolor{neutralcolor}{($-$)}}
& 68.4 {\scriptsize \textcolor{neutralcolor}{($-$)}}
& 41.6 {\scriptsize \textcolor{neutralcolor}{($-$)}}
& 73.2 {\scriptsize \textcolor{neutralcolor}{($-$)}}
& 33.8 {\scriptsize \textcolor{neutralcolor}{($-$)}}
& 21.6 {\scriptsize \textcolor{neutralcolor}{($-$)}}
& 56.6 {\scriptsize \textcolor{neutralcolor}{($-$)}}
& 46.9 {\scriptsize \textcolor{neutralcolor}{($-$)}}\\

\rowcolor{stageonebg}
\qquad+ \textcolor{stageone}{InternVideo2$_{\textit{stage1}}$}  
& \ding{51} 
& 98.1 {\scriptsize \textcolor{gaincolor}{(+47.4)}} 
& 74.4 {\scriptsize \textcolor{gaincolor}{(+6.0)}} 
& 42.8 {\scriptsize \textcolor{gaincolor}{(+1.2)}} 
& 73.9 {\scriptsize \textcolor{gaincolor}{(+0.7)}} 
& 36.3 {\scriptsize \textcolor{gaincolor}{(+2.5)}} 
& 28.9 {\scriptsize \textcolor{gaincolor}{(+7.3)}} 
& 56.2 {\scriptsize \textcolor{degradecolor}{(-0.4)}} 
& 45.9 {\scriptsize \textcolor{degradecolor}{(-1.0)}} \\

\rowcolor{stagetwobg}
\qquad+ \textcolor{stagetwo}{InternVideo2$_{\textit{stage2}}$}  
& \ding{55} 
& 41.0  {\scriptsize \textcolor{neutralcolor}{($-$)}}
& 71.7 {\scriptsize \textcolor{neutralcolor}{($-$)}}
& 40.7 {\scriptsize \textcolor{neutralcolor}{($-$)}}
& 72.3 {\scriptsize \textcolor{neutralcolor}{($-$)}}
& 33.8 {\scriptsize \textcolor{neutralcolor}{($-$)}}
& 22.4 {\scriptsize \textcolor{neutralcolor}{($-$)}}
& 56.0 {\scriptsize \textcolor{neutralcolor}{($-$)}}
& 45.0 {\scriptsize \textcolor{neutralcolor}{($-$)}}\\

\rowcolor{stagetwobg}
\qquad+ \textcolor{stagetwo}{InternVideo2$_{\textit{stage2}}$}  
& \ding{51} 
& 82.6 {\scriptsize \textcolor{gaincolor}{(+41.6)}} 
& 70.6 {\scriptsize \textcolor{degradecolor}{(-1.1)}} 
& 40.8 {\scriptsize \textcolor{gaincolor}{(+0.1)}} 
& 70.5 {\scriptsize \textcolor{degradecolor}{(-1.8)}} 
& 35.0 {\scriptsize \textcolor{gaincolor}{(+1.2)}} 
& 23.3 {\scriptsize \textcolor{gaincolor}{(+0.9)}} 
& 57.9 {\scriptsize \textcolor{gaincolor}{(+1.9)}} 
& 44.7 {\scriptsize \textcolor{degradecolor}{(-0.3)}} \\
\bottomrule
\end{tabular}
}
\end{table*}
In summary, temporal information is encoded in the early-to-intermediate layers of both $Stage1$ and $Stage2$, but $Stage1$ preserves this information more explicitly and transfers it effectively to the LLM. These results show that temporal information transfer requires both an appropriate projector and vision representations that retain explicit temporal information.

\section{RQ3: Can AoT supervision improve Video-LLM performance on broader temporal reasoning tasks?}
\label{section:rq3}

Based on the findings of RQ1 and RQ2, we adopt a Video-LLM configuration designed for effective temporal information transfer, combining temporal-aware video encoder and time-preserved projector with LLMs. We then investigate whether AoT supervision can improve Video-LLM performance on broader temporal reasoning tasks.

\subsection{Experiments of Instruction-Tuning}

\noindent\textbf{Model.} For the video encoder, we evaluate both InternVideo2$_{\textit{stage1}}$ and InternVideo2$_{\textit{stage2}}$.
We use the time-preserved MLP projector that reduces only the spatial resolution to $8 \times 8$ while keeping the temporal length unchanged at $T'=16$.
We use Mistral-7B-Instruct and Qwen2.5-7B-Instruct as the language models. During fine-tuning, the video encoder is kept frozen, the projector is trained from scratch, LLM is updated with LoRA.


\noindent\textbf{Instruction-Tuning Data.}
We fine-tune Video-LLMs using approximately 1 million general visual instruction-tuning samples \citep{maaz2024videogpt+}, which equip the models with basic visual understanding ability. This serves as the baseline training setting without AoT-specific supervision. Dataset details are provided in Appendix~\ref{sec:app_dataset}.

\noindent\textbf{AoT VQA Data.}
To study the effect of AoT supervision, we add AoT VQA samples to the instruction-tuning mixture. According to our additional probing results (Appendix~\ref{sec:app_layer}), we use videos with rich motion and physical dynamics, including SSv2 and Moments-in-Time (MiT) \citep{monfort2019momentstimedatasetmillion}.

\subsection{Benchmarks}



Following prior work on temporal evaluation in Video-LLMs \citep{xue2025seeing, cores2025tvbench, rasekh2026enhancing, li2024temporal}, we select benchmarks that differ in their sensitivity to temporal dynamics. We evaluate AoT supervision on temporal-sensitive benchmarks, while using general video benchmarks as temporal-neutral controls.

\noindent\textbf{Temporal Benchmarks.}
To cover diverse forms of temporal reasoning, we evaluate on AoT$_{PPB}$, TVBench \citep{cores2025tvbench}, and VITATECS (Direction) \citep{li2024vitatecsdiagnosticdatasettemporal}. These benchmarks require models to reason about physical irreversibility, temporal ordering, motion direction or temporal localization.

\noindent\textbf{General Benchmarks.}
We use MVBench \citep{li2024mvbenchcomprehensivemultimodalvideo} and Video-MME
\citep{fu2025videomme} as temporal-neutral controls. These benchmarks cover general video understanding tasks, domains, and video durations, but with low sensitivity to temporal information. We therefore use them as general-purpose control benchmarks to assess whether the additional AoT supervision will disrupt general video understanding.


\subsection{Main Results}

Table~\ref{tab:main} summarizes the main results. We include GPT-4o, Gemini2.5-pro, InternVideo2VL, Qwen2VL-7B and Qwen2.5VL-7B as reference baselines. The results on broader tasks are not directly comparable due to their substantially larger training data and different experimental settings.

\noindent\textbf{AoT supervision improves both AoT recognition and broader temporal reasoning.}
Adding AoT supervision yields substantial gains on AoT$_{PPB}$ across all configurations. 
Beyond AoT$_{PPB}$, it also improves performance on broader temporal reasoning benchmarks, with gains of up to 1.3 points on TVBench, and 6.0 points on VITATECS-Direction. Meanwhile, performance on MVBench and Video-MME remains stable, indicating that AoT supervision enhances temporal reasoning without degrading general video understanding.
On TVBench, the gains are concentrated in temporally sensitive subtasks, including Action Sequence (AS), Action Localization (AL), and Moving Direction (MD). Complete TVBench results are provided in Appendix~\ref{sec:app_tvbench}.

\noindent\textbf{Stage1 encoders are more effective for AoT supervision.}
Across both LLMs, InternVideo2$_{\textit{stage1}}$ consistently outperforms InternVideo2$_{\textit{stage2}}$ after AoT supervision and yields stronger gains on broader temporal reasoning tasks. 
This is consistent with our findings in RQ2: CLIP-style training makes temporal information less explicitly preserved and less transferable.





\begin{figure}[t]
    \centering
    \includegraphics[width=\linewidth]{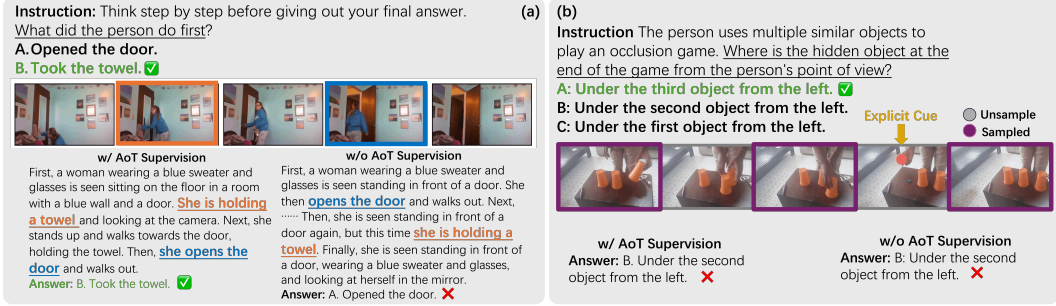}
    \caption{\textbf{(a) Case study.} The instruction requires the Video-LLM to identify the temporal order between \textcolor{orange}{taking the towel} and \textcolor{blue}{opening the door}. \textbf{(b) Error analysis.} Sparse frame sampling can miss critical evidence explicitly shown in the video.}
    \label{fig:case_study}
\end{figure}


\noindent\textbf{Case study and limitations.}
Figure~\ref{fig:case_study} presents case studies from TVBench. In Figure~\ref{fig:case_study}(a), AoT supervision helps the Video-LLM correctly identify that the person \textcolor{orange}{took the towel} before \textcolor{blue}{opening the door}, whereas the model without AoT supervision hallucinates an inconsistent temporal order. This suggests that AoT supervision improves reasoning over temporal dynamics and event progression. 

Figure~\ref{fig:case_study}(b) shows an error case caused by a limited number of frames. We use only 16 input frames throughout our experiments, due to limited computational resources, which can miss explicit evidence and cause the model to predict incorrectly. 
This error case indicates that datasets like TVBench are not only limited by temporal reasoning, but are also affected by other factors, such as frame number setting, which potentially confound observations for the true impact of AoT supervision.


\section{Conclusion}

We study why Video-LLMs struggle with the Arrow-of-Time task by tracing how temporal information flows through vision encoders, projectors, and LLMs. Our analysis shows that video-centric models encode strong AoT-relevant information, but this information is often lost during transfer to the LLM. Guided by these findings, we build a Video-LLM that surpasses human performance on the AoT$_{PPB}$ and show that AoT supervision further benefits broader temporal reasoning benchmarks. Overall, our work identifies the key bottlenecks in Video-LLMs for temporal information and offers practical principles for improving temporal reasoning capabilities.
\small
\bibliographystyle{plainnat}
\bibliography{custom}


\appendix
\section{InternVideo2 Details}
\label{sec:app_internvideo}

\noindent\textbf{InternVideo2$_{\text{stage1}}$: Vision Encoder Pre-training.}
InternVideo2$_{\text{stage1}}$ \citep{wang2024internvideo2} is trained as a video-centric encoder through masked video reconstruction. 
Given uniformly sampled video frames, a large portion of video patches is masked, and the encoder is optimized to reconstruct the video representations distilled from video expert models \citep{tong2022videomae, chen2024internvlscalingvisionfoundation}. 
This stage mainly aims to learn spatiotemporal representations from large-scale video data, providing the basis for subsequent multimodal alignment.

\noindent\textbf{InternVideo2$_{\text{stage2}}$: Language Alignment for Vision Encoder.}
InternVideo2$_{\text{stage2}}$ further aligns the InternVideo2$_{\text{stage1}}$ with language by training on video--text and image--text data. 
The model is primarily optimized with CLIP-style training objective, enabling the visual representations to be aligned with language. 
Compared with $stage1$, this stage emphasizes language alignment rather than purely video-centric representation learning.

\noindent\textbf{Instruction Tuning (InternVideo2VL).}
InternVideo2VL couples InternVideo2$_{\text{stage2}}$ with Mistral-7B via a Q-Former projector and is trained on multimodal instruction-following data. 
The video encoder and Q-Former are fully trainable, while the LLM is adapted with LoRA. 
This stage adapts the model for conversation and question answering, allowing the LLM to generate language responses conditioned on video representations.

\section{Additional Experiments}
\label{sec:app_experiments}

\noindent\textbf{Probing setup.}
For attentive probing, we train the probe for 1 epoch with a batch size of 4 and a learning rate of $1\times10^{-4}$. The attention-based pooling layer consists of one cross-attention layer with 16 attention heads and one additional learnable query token. The probe classifier is optimized with a standard cross-entropy loss. Probing dataset statistics is summarized in Table~\ref{table:data_stat}.

\noindent\textbf{Video-LLM fine-tuning setup.}
For Video-LLM fine-tuning, we keep the video encoder frozen, train the projector from scratch, and update the LLM with LoRA. We use LoRA rank 16 and $\alpha=32$. Models are trained for 1 epoch with a batch size of 2, a learning rate of $2\times10^{-5}$, and a weight decay of 0.02.

\noindent\textbf{Computational resource.} We conduct our experiments with 8 A100 80G GPUs.

\section{Additional Ablations}

\subsection{Layer-Wise Probing on Various Settings}
\label{sec:app_layer}

\noindent\textbf{Study on probe training datasets.}
To examine whether the layer-wise AoT dynamics are robust to the choice of probe training data, we train attentive probes on three video datasets: SSv2, Moments-in-Time (MiT), and K400 \citep{kay2017kineticshumanactionvideo}. For a fair comparison, we sample MiT$_{170k}$ and K400$_{170k}$ subsets from MiT and K400, respectively, matching the scale of SSv2 as shown in Table~\ref{table:data_stat}. We uniformly sample videos from each action class to construct these subsets. These datasets differ in average video duration and action distribution. For all settings, we construct balanced forward/backward supervision by randomly reversing 50\% of the training videos, and evaluate the trained probes on AoT$_{PPB}$.

\begin{figure*}[t]
    \centering
    \includegraphics[width=\linewidth]{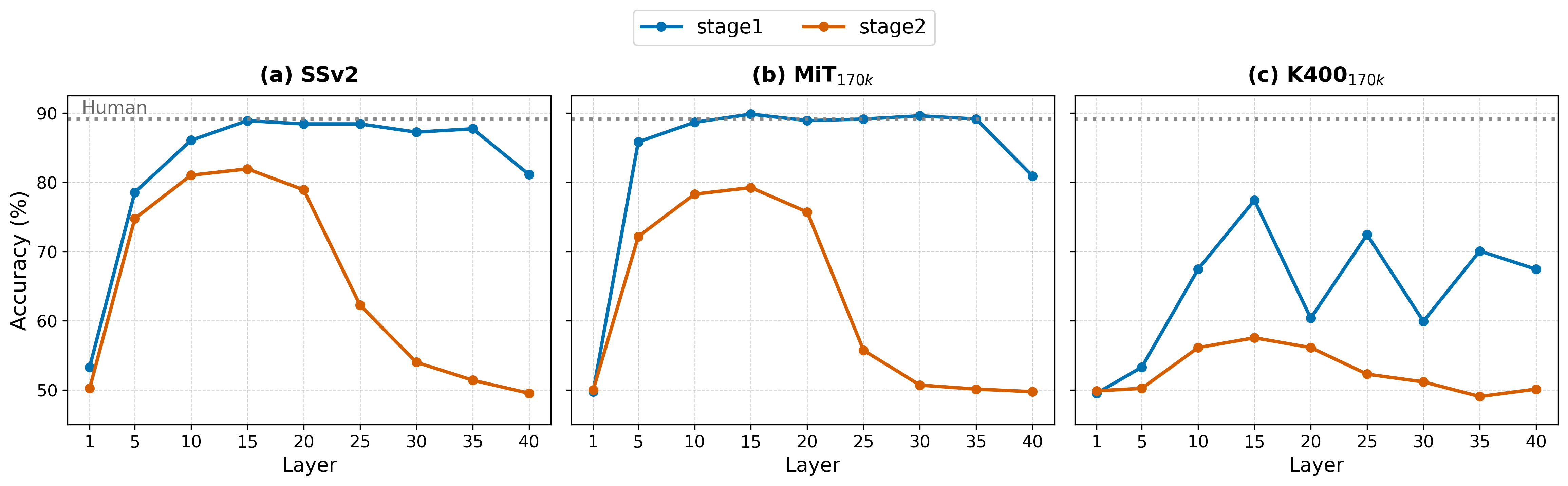}
    \caption{
    \textbf{Layer-wise AoT probing across probe training datasets.}
    Probes are trained on SSv2, MiT$_{170k}$, and K400$_{170k}$, and evaluated on AoT$_{PPB}$.
    InternVideo2$_{\textit{stage1}}$ maintains strong AoT performance across deeper layers, whereas InternVideo2$_{\textit{stage2}}$ peaks at early layers and then drops sharply.
    }
    \label{fig:layer_stage}
\end{figure*}

Figure~\ref{fig:layer_stage} shows that the overall trend is consistent across datasets. For SSv2 and MiT$_{170k}$, temporal information is encoded in the intermediate layers of InternVideo2$_{\textit{stage1}}$ and InternVideo2$_{\textit{stage2}}$, while the language-aligned stage shows clear degradation in later layers. K400$_{170k}$ yields weaker and less stable probing performance, likely because its videos are longer than those in SSv2 and MiT, causing uniform frame sampling to lose more temporal information. Nevertheless, the same qualitative pattern remains: InternVideo2$_{\textit{stage1}}$ preserves stronger temporal information than language-aligned InternVideo2$_{\textit{stage2}}$.


\begin{table}
\small
  \centering
  \caption{\label{table:data_stat}
   \textbf{Statistics of AoT Probing Datasets.}
  }
  \begin{tabular}{lccc}

   \textbf{Dataset} & \textbf{\# Duration} & \textbf{\# Action} & \textbf{\# Size} \\
    \hline
    \multicolumn{4}{c}{\textbf{Training}}  \\
    \hline
    $SSv2$ &4.03s &172 & 169k   \\
    $ MiT_{170k} $  &3.00s &339 & 170k  \\
    $ K400_{170k} $  &9.57s &400 & 170k \\
    \hline
    \multicolumn{4}{c}{\textbf{Evaluation}} \\
    \hline
    $ AoT_{PPB} $ & 3.00s & -  & 424 \\
    \bottomrule
  \end{tabular}
  
\end{table}

\begin{figure}[t]
    \centering
    \includegraphics[width=0.6\linewidth]{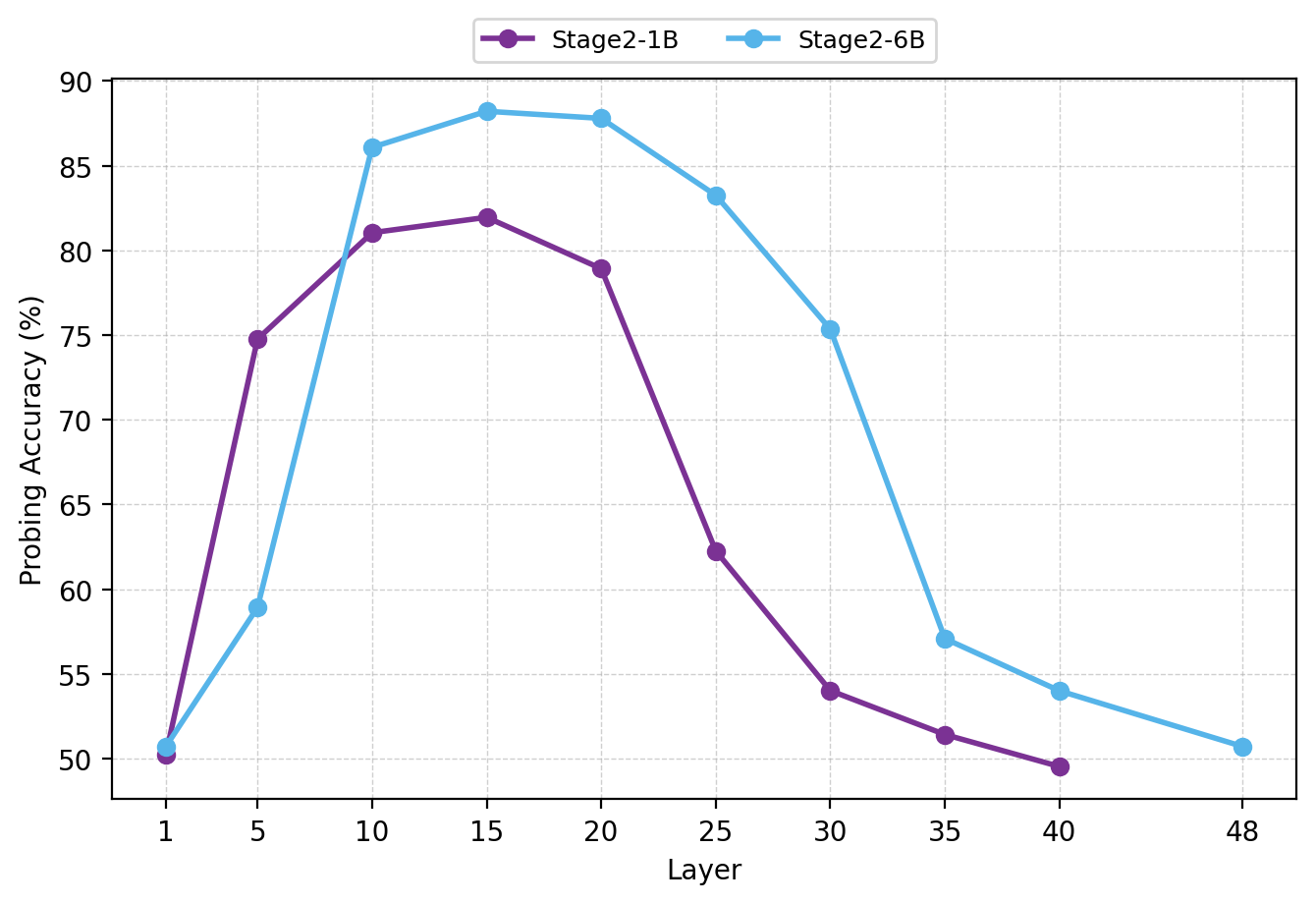}
    \caption{\textbf{Layer-wise probing performance of InternVideo2$_{stage2}$ 1B and 6B.} Probes are trained on SSv2.}
    \label{fig:layer_size}
\end{figure}

\noindent\textbf{Study on video-centric encoder size.}
We further examine whether the layer-wise AoT performance is affected by encoder size. Figure~\ref{fig:layer_size} compares InternVideo2$_{\textit{stage2}}$ with 1B and 6B parameters, using probes trained on SSv2. Both models exhibit the same overall trend: temporal information is encoded in the early layers and then degrades in later layers. The 6B encoder achieves higher peak probing accuracy and maintains strong performance over more layers, suggesting that larger video-centric encoders encode stronger AoT-relevant temporal information. However, its performance still drops toward chance level in the late layers, indicating that increasing encoder size does not eliminate the degradation of explicitly accessible temporal information after language alignment.

\subsection{Projector Analysis}
\label{sec:app_proj}

\begin{figure}[t]
    \centering
    \includegraphics[width=0.85\linewidth]{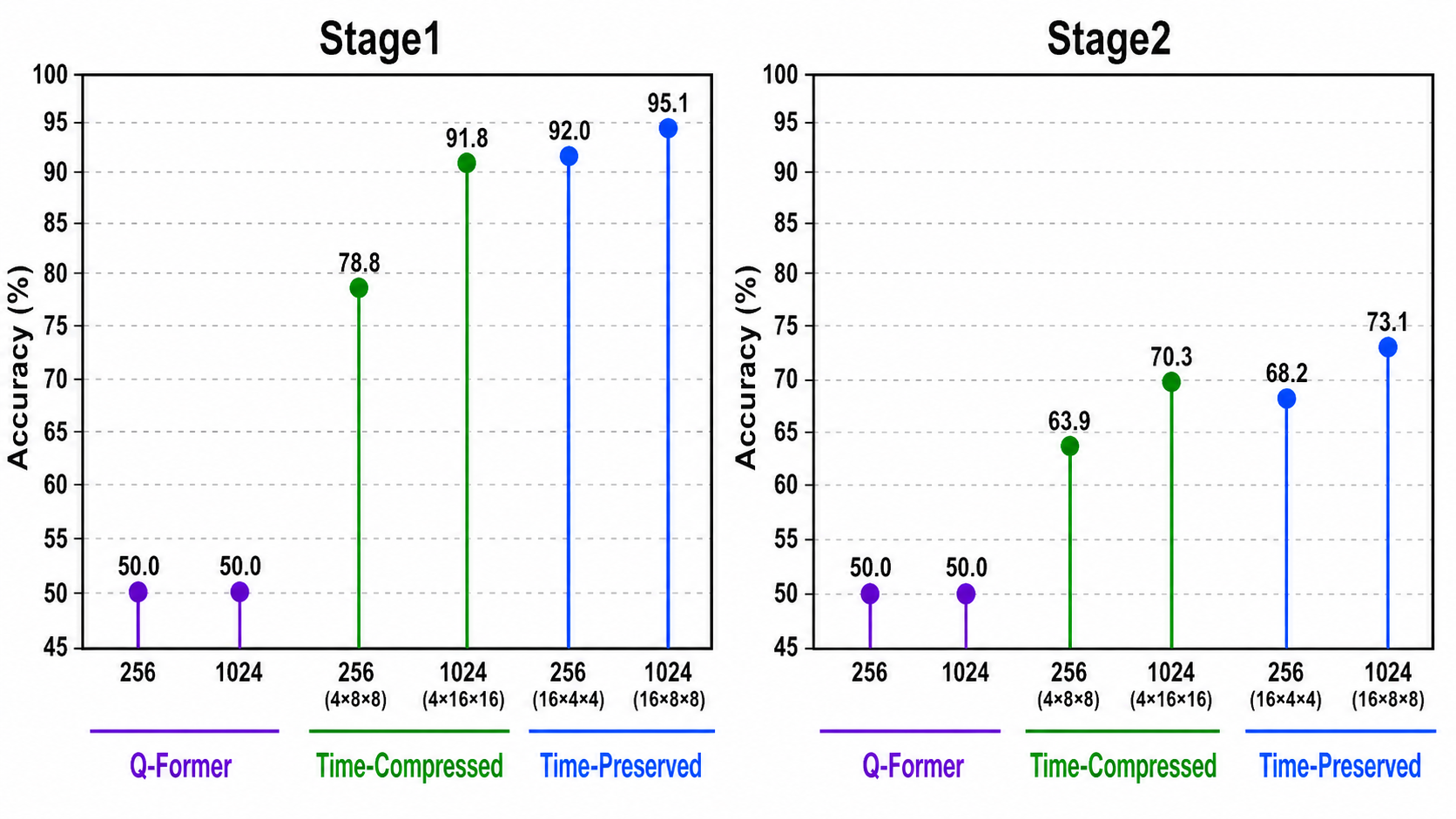}
    \caption{\textbf{Projector Ablation on AoT$_{PPB}$.} We compare MLP-based projectors with different pooling strategies and Q-Former projectors under two visual-token budgets. We report overall accuracy on AoT$_{PPB}$. Stage1 and Stage2 denote InternVideo2$_{\textit{stage1}}$ and InternVideo2$_{\textit{stage2}}$.}
    \label{fig:full_proj_ablation}
\end{figure}

Figure~\ref{fig:full_proj_ablation} compares projector designs on AoT$_{PPB}$ under two visual-token budgets. 
Q-Former projectors remain at chance level for both InternVideo2$_{\textit{stage1}}$ and InternVideo2$_{\textit{stage2}}$, regardless of whether 256 or 1024 query tokens are used. 
This suggests that query-based projection is ineffective for transferring temporal information.

In contrast, MLP-based projectors substantially improve AoT performance. 
For InternVideo2$_{\textit{stage1}}$, the time-compressed MLP reaches 78.8\% with 256 tokens and 91.8\% with 1024 tokens, while the time-preserved MLP further improves performance to 92.0\% and 95.1\%, respectively. 
For InternVideo2$_{\textit{stage2}}$, the same trend holds: time-compressed projection improves accuracy to 63.9\% and 70.3\%, while time-preserved projection achieves 68.2\% and 73.1\%. 
These results show that preserving temporal structure during projection is critical for effective information transfer.

Increasing the visual-token budget also consistently improves performance for MLP-based projectors, indicating that retaining more visual information benefits temporal reasoning. 
However, even with the same projector and token budget, InternVideo2$_{\textit{stage1}}$ consistently outperforms InternVideo2$_{\textit{stage2}}$, suggesting that language alignment weakens the transferability of AoT-relevant temporal information. 
Overall, the projector analysis identifies temporal preservation as a key requirement for transferring temporal information from video-centric encoders to LLMs.

\begin{table}[t]
\centering
\caption{\textbf{AoT VQA templates.}}
\label{tab:aot_vqa_templates}
\small
\begin{tabular}{p{0.95\linewidth}}
\toprule
\textbf{Template 1} \\
\textit{Instruction:} Examine the motion and temporal cues in the video to determine whether it is forward or backward. \\
\textit{Question:} Based on the temporal cues, is the video forward or backward? \\
\textit{Answer:} forward / backward \\
\midrule
\textbf{Template 2} \\
\textit{Instruction:} Determine whether the video is played normally or backward based on physical and temporal consistency. \\
\textit{Question:} Is this clip more consistent with real-world forward dynamics or backward playback? \\
\textit{Answer:} forward / backward \\
\midrule
\textbf{Template 3} \\
\textit{Instruction:} Identify the arrow of time in the video by deciding whether it runs forward or backward. \\
\textit{Question:} What is the arrow of time of this video? \\
\textit{Answer:} forward / backward \\
\midrule
\textbf{Template 4} \\
\textit{Instruction:} Analyze the temporal progression of the video and classify it as forward-time or reversed-time. \\
\textit{Question:} Which of the following best describes the temporal direction of this clip? \\
\textit{Answer:} forward / backward \\
\bottomrule
\end{tabular}
\end{table}

\section{Video-LLM Tuning Dataset}
\label{sec:app_dataset}

\subsection{Instruction Tuning Datasets}
Following \citet{maaz2024videogpt+}, we use approximately one million visual instruction-tuning samples, comprising both video--text and image--text instruction data.
The dataset is constructed from a diverse mixture of multimodal instruction data, including VideoChat \citep{li2024videochatchatcentricvideounderstanding}, K710 \citep{kay2017kineticshumanactionvideo}, SSv2 \citep{goyal2017something}, VideoChatGPT \citep{Maaz2023VideoChatGPT}, CLEVRER \citep{yi2020clevrercollisioneventsvideo}, Next-QA \citep{xiao2021nextqa}, VCG-112k \citep{maaz2024videogpt+}, VCG-human \citep{maaz2024videogpt+}, WebVid \citep{Bain21}, ActivityNet \citep{Heilbron_2015_CVPR}, COCO \citep{lin2015microsoftcococommonobjects}.
These data cover diverse instruction formats, including action classification, temporal reasoning, image/video captioning, image/video question answering, and video conversation, providing broad supervision for aligning visual representations with the language model.

\newcommand{\gcell}[2]{\cellcolor{gaincolor!#1}#2}
\newcommand{\dcell}[2]{\cellcolor{degradecolor!#1}#2}
\newcommand{\ncell}[1]{\cellcolor{neutralcolor!10}#1}
\begin{table*}[t]
\centering
\caption{\textbf{Categorical Results on TVBench.} The $+$AoT column indicates whether AoT data is included during instruction tuning. Colored deltas report performance changes relative to the corresponding counterpart trained without AoT. 
Baseline models are not directly comparable to ours due to substantially larger training data and different experimental settings.
}
\label{tab:tvbench}
\setlength{\tabcolsep}{3pt}
\resizebox{\textwidth}{!}{
\begin{tabular}{lcccccccccccc}
\toprule
\multirow{2}{*}{\textbf{Model}} & \multirow{2}{*}{\textbf{$+$AoT}}  & \multirow{2}{*}{\makecell{\textbf{TVBench}\\\textbf{Averaged}}} & \multicolumn{10}{c}{\textbf{TVBench}} \\
\cmidrule(lr){4-13}
 &  &  & AC & OC & AS & OS & ST & AL & AA & UA & ES & MD \\
\midrule
GPT-4o &  & 39.9 & 26.1 & 21.3 & 59.3 & 33.2 & 52.4 & 25.0 & 78.4 & 41.7 & 31.0 & 30.6 \\
\midrule
InternVideo2VL & & 40.7 & 28.2 & 46.0 & 65.2 & 33.8 & 63.8 & 38.8 & 57.8 & 29.2 & 17.5 & 26.7 \\
Qwen2VL-7B & & 43.8 & 27.1 & 61.5 & 63.8 & 38.7 & 75.7 & 41.3 & 63.1 & 22.0 & 22.5 & 22.4 \\
Qwen2.5VL-7B & & 45.2 & 36.8 & 36.5 & 64.8 & 37.3 & 62.7 & 38.8 & 74.1 & 41.5 & 26.0 & 33.6 \\
\midrule

\multicolumn{13}{l}{\textbf{Mistral-7B}} \\

\rowcolor{stageonebg}
\qquad+ \textcolor{stageone}{InternVideo2$_{\textit{stage1}}$} 
& \ding{55} 
& 40.1 {\scriptsize \textcolor{neutralcolor}{($-$)}}  
& 25.0 & 39.8 & 64.6 & 38.7 & 69.7 & 34.4 & 52.2 & 36.6 & 14.5 & 25.4 \\

\rowcolor{stageonebg}
\qquad+ \textcolor{stageone}{InternVideo2$_{\textit{stage1}}$} 
& \ding{51} 
& 41.4 {\scriptsize \textcolor{gaincolor}{(+1.3)}}  
& \ncell{25.0} 
& \dcell{25}{37.8} 
& \gcell{35}{68.7} 
& \dcell{25}{35.1} 
& \dcell{15}{68.1} 
& \gcell{25}{38.1} 
& \gcell{25}{55.3} 
& \ncell{36.6} 
& \gcell{35}{19.5} 
& \gcell{35}{29.7} \\

\rowcolor{stagetwobg}
\qquad+ \textcolor{stagetwo}{InternVideo2$_{\textit{stage2}}$} 
& \ding{55}
& 39.0 {\scriptsize \textcolor{neutralcolor}{($-$)}} 
& 26.9 & 41.9 & 64.9 & 34.7 & 69.8 & 29.4 & 49.1 & 32.9 & 11.5 & 29.3 \\

\rowcolor{stagetwobg}
\qquad+ \textcolor{stagetwo}{InternVideo2$_{\textit{stage2}}$} 
& \ding{51}
& 39.6 {\scriptsize \textcolor{gaincolor}{(+0.6)}}  
& \dcell{15}{25.4} 
& \dcell{35}{37.2} 
& \gcell{25}{67.3} 
& \dcell{25}{32.4} 
& \gcell{15}{71.4} 
& \gcell{15}{31.3} 
& \gcell{25}{52.2} 
& \gcell{15}{34.1} 
& \gcell{45}{19.0} 
& \dcell{25}{25.4} \\

\midrule

\multicolumn{13}{l}{\textbf{Qwen2.5-7B}} \\

\rowcolor{stageonebg}
\qquad+ \textcolor{stageone}{InternVideo2$_{\textit{stage1}}$} 
& \ding{55}
& 41.6 {\scriptsize \textcolor{neutralcolor}{($-$)}}  
& 25.1 & 43.2 & 73.2 & 34.2 & 77.3 & 33.8 & 55.6 & 32.9 & 18.5 & 21.6 \\

\rowcolor{stageonebg}
\qquad+ \textcolor{stageone}{InternVideo2$_{\textit{stage1}}$} 
& \ding{51}
& 42.8 {\scriptsize \textcolor{gaincolor}{(+1.2)}}  
& \gcell{25}{27.1} 
& \gcell{35}{48.6} 
& \gcell{10}{73.9} 
& \ncell{34.2} 
& \dcell{25}{75.1} 
& \gcell{25}{36.3} 
& \dcell{15}{53.8} 
& \dcell{15}{31.7} 
& \dcell{10}{18.0} 
& \gcell{45}{28.9} \\

\rowcolor{stagetwobg}
\qquad+ \textcolor{stagetwo}{InternVideo2$_{\textit{stage2}}$} 
& \ding{55}
& 40.7 {\scriptsize \textcolor{neutralcolor}{($-$)}} 
& 25.4 & 37.8 & 72.3 & 38.7 & 70.8 & 33.8 & 56.9 & 32.9 & 16.0 & 22.4 \\

\rowcolor{stagetwobg}
\qquad+ \textcolor{stagetwo}{InternVideo2$_{\textit{stage2}}$} 
& \ding{51}
& 40.8 {\scriptsize \textcolor{gaincolor}{(+0.1)}}  
& \gcell{10}{25.9} 
& \gcell{25}{40.5} 
& \dcell{15}{70.5} 
& \dcell{25}{35.1} 
& \dcell{45}{64.3} 
& \gcell{15}{35.0} 
& \gcell{10}{57.5} 
& \ncell{32.9} 
& \gcell{45}{22.5} 
& \gcell{10}{23.3} \\

\bottomrule
\end{tabular}%
}
\end{table*}

\subsection{AoT VQA Datasets}
\label{sec:aot_vqa_templates}

We combine the SSv2 training split and MiT$_{170k}$ (Table~\ref{table:data_stat}). To convert forward/backward videos into VQA-style samples, we manually design four instruction templates. We use multiple templates to reduce instruction-specific bias and prevent the model from overfitting to a fixed question pattern. For each video, we uniformly select one template and assign the answer according to its playback direction, i.e., \texttt{forward} or \texttt{backward}. The templates are shown in Table~\ref{tab:aot_vqa_templates}.

\section{Categorical Results on TVBench}
\label{sec:app_tvbench}

To better understand how AoT supervision affects different types of temporal reasoning, we report category-level results on TVBench in Table~\ref{tab:tvbench}. Overall, adding AoT supervision yields consistent improvements in the averaged TVBench score. With Mistral-7B, AoT supervision improves InternVideo2$_{\textit{stage1}}$ from 40.1 to 41.4 and InternVideo2$_{\textit{stage2}}$ from 39.0 to 39.6. With Qwen2.5-7B, it improves InternVideo2$_{\textit{stage1}}$ from 41.6 to 42.8 and InternVideo2$_{\textit{stage2}}$ from 40.7 to 40.8.

At the category level, the gains are not uniform. AoT supervision tends to improve categories that require recognizing motion or temporal changes. For example, with Mistral-7B and InternVideo2$_{\textit{stage1}}$, AoT supervision improves Action
Sequence (AS) from 64.6 to 68.7, Action
Antonym (AA) from 52.2 to 55.3, and Egocentric
Sequence (ES) from 14.5 to 19.5.  In contrast, categories such as Object Shuffle (OS), which rely more on object perception, show smaller or mixed changes. This suggests that AoT supervision mainly strengthens temporal sensitivity.

These results indicate that AoT supervision provides a useful but targeted temporal training signal. It improves TVBench on average, but the gains are smaller than those on AoT$_{PPB}$ and VITATECS-Direction, likely because TVBench covers more diverse video durations and our models use a fixed 16-frame input.

\end{document}